\PassOptionsToPackage{hyphens}{url}
\documentclass{article}

\usepackage[final,eandd]{neurips_2026}

\usepackage[utf8]{inputenc}
\usepackage[T1]{fontenc}
\usepackage{hyperref}
\usepackage{url}
\usepackage{booktabs}
\usepackage{tabularx}
\usepackage{amsfonts}
\usepackage{amsmath}
\usepackage{amssymb}
\usepackage{nicefrac}
\usepackage[expansion=false]{microtype}
\usepackage{xcolor}
\usepackage{graphicx}
\usepackage{float}
\usepackage{subcaption}
\usepackage{multirow}
\usepackage{enumitem}
\usepackage{siunitx}
\newcommand{\TableStyle}{%
  \small
  \setlength{\tabcolsep}{5pt}%
  \renewcommand{\arraystretch}{1.1}%
}

\newcommand{\NUMBLOCKS}{112}
\newcommand{\NUMCITIESVTHREE}{25}
\newcommand{\NUMIMAGESVTHREE}{7563}
\newcommand{\NUMAVGBLOCK}{67.5}

\newcommand{\CROSSCITYMEAN}{0.284}

\newcommand{\GEOVISSPEARMAN}{0.180}
\newcommand{\GEOVISPVAL}{1.8 \times 10^{-3}}

\newcommand{\CITYTABLEVTHREE}{Amsterdam & NL & 52.37 & 4.90 & R & 6 & 398 \\
Bangkok & TH & 13.76 & 100.50 & L & 3 & 170 \\
Berlin & DE & 52.52 & 13.40 & R & 1 & 28 \\
Bogota & CO & 4.71 & $-$74.07 & R & 2 & 153 \\
Buenos Aires & AR & $-$34.60 & $-$58.38 & R & 6 & 376 \\
Cairo & EG & 30.04 & 31.24 & R & 2 & 100 \\
Cape Town & ZA & $-$33.92 & 18.42 & L & 3 & 189 \\
Dubai & AE & 25.20 & 55.27 & R & 7 & 445 \\
Istanbul & TR & 41.01 & 28.98 & R & 5 & 300 \\
London & GB & 51.51 & $-$0.13 & L & 7 & 445 \\
Melbourne & AU & $-$37.81 & 144.96 & L & 1 & 68 \\
Mexico City & MX & 19.43 & $-$99.13 & R & 2 & 151 \\
Mumbai & IN & 19.08 & 72.88 & L & 6 & 403 \\
Nairobi & KE & $-$1.29 & 36.82 & L & 1 & 38 \\
New York & US & 40.71 & $-$74.01 & R & 9 & 767 \\
Paris & FR & 48.86 & 2.35 & R & 8 & 500 \\
Rome & IT & 41.90 & 12.50 & R & 5 & 382 \\
San Francisco & US & 37.77 & $-$122.42 & R & 2 & 66 \\
Sao Paulo & BR & $-$23.55 & $-$46.63 & R & 3 & 142 \\
Seoul & KR & 37.57 & 126.98 & R & 5 & 311 \\
Shanghai & CN & 31.23 & 121.47 & R & 4 & 334 \\
Singapore & SG & 1.35 & 103.82 & L & 3 & 182 \\
Sydney & AU & $-$33.87 & 151.21 & L & 4 & 213 \\
Tokyo & JP & 35.68 & 139.65 & L & 10 & 910 \\
Toronto & CA & 43.65 & $-$79.38 & R & 7 & 492}
\newcommand{\VThreeDeltaLOneLZeroC}{0.036}
\newcommand{\VThreeDeltaLTwoLOneC}{0.004}

\newcommand{\VThreeLZeroSDXLC}{0.512}
\newcommand{\VThreeLOneSDXLC}{0.536}
\newcommand{\VThreeLTwoSDXLC}{0.536}
\newcommand{\VThreeLZeroSDLC}{0.432}
\newcommand{\VThreeLOneSDLC}{0.515}
\newcommand{\VThreeLTwoSDLC}{0.516}
\newcommand{\VThreeLZeroFLDC}{0.456}
\newcommand{\VThreeLOneFLDC}{0.484}
\newcommand{\VThreeLTwoFLDC}{0.496}
\newcommand{\VThreeLZeroFLSC}{0.465}
\newcommand{\VThreeLOneFLSC}{0.499}
\newcommand{\VThreeLTwoFLSC}{0.507}
\newcommand{\VThreeLZeroPXC}{0.459}
\newcommand{\VThreeLOnePXC}{0.486}
\newcommand{\VThreeLTwoPXC}{0.489}
\newcommand{\VThreeLZeroHDC}{0.458}
\newcommand{\VThreeLOneHDC}{0.477}
\newcommand{\VThreeLTwoHDC}{0.478}

\title{GeoFidelity-Bench: Evaluating Block-Conditioned Geographic Fidelity in Text-to-Image Street-View Generation}

\author{
  Kaizhen Tan\textsuperscript{1,2,3}\\
  \textsuperscript{1}Heinz College of Information Systems and Public Policy, Carnegie Mellon University\\
  \textsuperscript{2}Robert F. Wagner Graduate School of Public Service, New York University\\
  \textsuperscript{3}Shanghai Key Laboratory of Urban Design and Urban Science, NYU Shanghai\\
  \texttt{kt3275@nyu.edu}
}

\begin{document}

\maketitle

\begin{abstract}
Text-to-image models can generate plausible street scenes from prompts
such as ``a street in New York City,'' but it remains unclear whether these
images resemble the requested place or show a generic city appearance.
Commonly used evaluation metrics mainly measure image quality or text
alignment and do not directly assess geographic fidelity. We
introduce \textbf{GeoFidelity-Bench}, a benchmark for evaluating whether
generated street-view images match a target location at the level of
named street blocks. The benchmark contains \num{\NUMBLOCKS} street
blocks from \num{\NUMCITIESVTHREE} cities across six continents, each paired
with a panel of real Mapillary images and geographic metadata. From this
metadata we build prompts with the city only, with street and
neighborhood names, and with these names plus raw GPS coordinates.
Across six open-weight text-to-image models, street and neighborhood
names improve similarity to the target location, while raw GPS
coordinates add only a small and uneven gain. Control prompts that swap
in the names of other blocks while keeping the city fixed reduce
retrieval accuracy, with larger decreases when the neighborhood name or
both names are replaced. Generated images also distinguish their target
blocks from local alternatives less reliably than held-out real images.
CLIP alignment between generated images and their prompts is nearly
uninformative about whether the images achieve block-level geographic
fidelity. GeoFidelity-Bench provides a controlled testbed for studying
location-conditioned generation and for measuring progress beyond generic
city-level plausibility.
\end{abstract}

\section{Introduction}
\label{sec:intro}

Modern text-to-image models can generate realistic street scenes from place
names. A model may produce a believable ``street in New York City'' while
relying on a broad city stereotype.
Fr\'echet Inception Distance (FID)~\citep{heusel2017fid},
CLIPScore~\citep{hessel2021clipscore}, and learned perceptual image patch
similarity (LPIPS)~\citep{zhang2018lpips}
measure distributional similarity, text--image alignment, and perceptual
image distance. Without real images of the requested location and of
nearby alternatives, these scores cannot show whether a generated image
resembles that place.

Location-conditioned generation is becoming more common. Recent work
conditions generation on GPS coordinates~\citep{feng2025gps}, structured
street representations~\citep{deng2024streetscapes}, and satellite
imagery~\citep{li2024sat2scene,xu2024geospecific}.
\citet{feng2025gps} score the compatibility between a generated image and
its input coordinates, and \citet{jang2024place} evaluate place identity
at city scale. Researchers developing these systems also need to know
whether a generated image matches the requested street better than
visually plausible streets in the same city. This test requires real
references tied to individual locations and an evaluation protocol that
compares each location with its neighbors.

We evaluate at the level of named street blocks, which are segments of
named streets in OpenStreetMap (OSM)~\citep{openstreetmap}. Reference
images for a block must account for variation within it, because a change
in camera position or heading can alter which facades, road surfaces, and
vegetation are visible. We therefore represent each block with a
reference panel, a set of real images covering several views. Comparing
a generated image with the panels of other blocks in the same city then
tests whether it favors the requested street over locally plausible
alternatives.

GeoFidelity-Bench contains \num{\NUMBLOCKS} such blocks in
\num{\NUMCITIESVTHREE} cities, each paired with a panel of real Mapillary
images. We evaluate six text-to-image models under city-only prompts,
prompts containing street and neighborhood names, and prompts that
additionally include GPS coordinates. Generated images are compared with
the target reference panel and with the panels of alternative blocks.

Adding street and neighborhood names improves similarity to the target
reference panel for all six generators, while appending raw GPS
coordinates produces a much smaller gain. Even with these names,
generated images rank their target block first among the candidate panels
in 40\% of cases on average, compared with 86\% for held-out real images.
CLIP alignment with the generation prompt is nearly unrelated to this
retrieval outcome.

\paragraph{Contributions.}
\begin{enumerate}[leftmargin=*, itemsep=2pt, topsep=2pt]
    \item We build \textbf{GeoFidelity-Bench}, a street-view benchmark of
          \num{\NUMBLOCKS} named blocks in \num{\NUMCITIESVTHREE} cities,
          each with a Mapillary reference panel (\num{\NUMIMAGESVTHREE}
          references in total).
    \item We define a \textbf{prompt-conditioning study} that adds city
          names, street and neighborhood names, and GPS coordinates to the
          prompt, with controls that replace the street or neighborhood
          name while keeping the city fixed.
    \item We provide \textbf{geographic fidelity diagnostics} for
          target-panel similarity, set-level fidelity and diversity,
          semantic composition, and discrimination from geographically
          related alternatives, validated with held-out real images.
    \item We report a \textbf{six-model evaluation} with held-out
          real-image anchors, same-city control prompts, and a comparison
          with CLIP text alignment.
\end{enumerate}

\section{Related Work}
\label{sec:related}

\paragraph{Location-conditioned generation.}
Generation methods differ in the geographic information they receive.
\citet{feng2025gps} use GPS coordinates and text, while
Streetscapes~\citep{deng2024streetscapes} uses structured urban inputs
to synthesize street-view sequences. Cross-view methods use satellite
imagery to guide ground-level scenes~\citep{li2024sat2scene,xu2024geospecific}.
At a larger spatial scale, CityDreamer~\citep{xie2024citydreamer} and
UrbanWorld~\citep{shang2024urbanworld} generate 3D urban environments.
These conditioning formats supply different spatial cues; evaluating
the resulting appearance requires references tied to the requested place.

\paragraph{Geographic recognition and retrieval.}
GeoCLIP~\citep{vivanco2023geoclip}, PIGEON~\citep{haas2024pigeon}, and
OpenStreetView-5M~\citep{astruc2024osv5m} study how to infer location
from images. Visual place recognition formulates a related task as
image retrieval, with methods such as MixVPR~\citep{alibey2023mixvpr}
and AnyLoc~\citep{keetha2023anyloc} matching queries to reference images.
This matching formulation motivates a local test of whether a generated
image resembles its target block more than other blocks from the same city.

\paragraph{Geographic representation and place identity.}
GeoDE~\citep{ramaswamy2023geode} provides geographically diverse images
for evaluating object recognition. For generated images, DIG
In~\citep{hall2023dig} measures regional disparities in quality and
diversity; \citet{sureddy2024decomposed} separate these comparisons into
objects and backgrounds. At city scale, \citet{jang2024place} evaluate
place identity in DALL-E~2 outputs using web references and human
judgments. Our comparison uses named street blocks as the target unit
and real street-view panels as references, allowing comparisons within
the same city.

\paragraph{Text-to-image evaluation.}
Prompt-alignment metrics differ in how they check the depicted content.
CLIPScore~\citep{hessel2021clipscore} summarizes image--text compatibility
in a shared embedding space. More detailed checks use object detection
for count, position, and color in GenEval~\citep{ghosh2023geneval}, or
questions derived from the prompt in TIFA~\citep{hu2023tifa}.
These checks evaluate information expressed in the prompt. A street
name leaves much of the expected appearance implicit, so we compare
generated images with real images of the named location.
Section~\ref{sec:text_alignment} measures how CLIP alignment relates to
this comparison.

\section{GeoFidelity-Bench}
\label{sec:benchmark}

\begin{figure}[H]
\centering
\includegraphics[width=\textwidth]{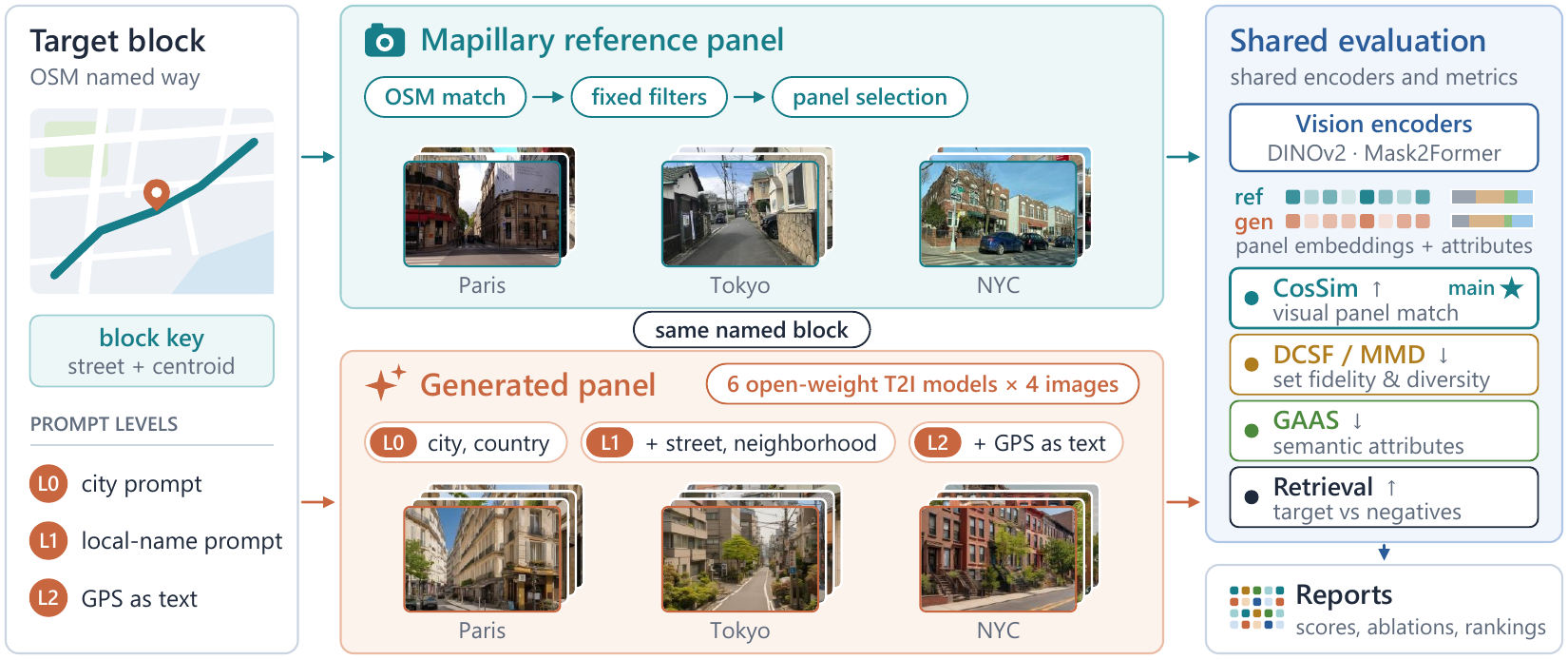}
\caption{\textbf{Overview of GeoFidelity-Bench.} Each target is a named
OSM street block. Its geometry selects the Mapillary images that form the
reference panel, and its metadata fills three prompt levels: city only
(L0), street and neighborhood names (L1), and L1 plus GPS coordinates as
text (L2). Six open-weight text-to-image models each generate four images
per block and prompt level. Shared vision encoders compare the generated
and reference panels through visual similarity, set fidelity and
diversity, semantic attribute agreement, and retrieval against the panels
of related blocks. Thumbnails show real references and SDXL images under
L1 prompts for one block in each of Paris, Tokyo, and New York.}
\label{fig:framework}
\end{figure}

Each target is a named street block with a real-image reference panel
and geographic metadata. We vary the location information supplied to
the generator, then compare its outputs with the target panel and
geographically related alternatives (Figure~\ref{fig:framework}).

\subsection{Target Blocks and Reference Collection}

We start from \num{\NUMCITIESVTHREE} cities on six continents. For each
city, we query OSM for named \texttt{highway=*} ways within 6\,km of the
city center. We use \emph{block} to denote a named OSM way or a segment
carved from it; this unit need not coincide with an intersection-bounded
city block. We stratify ways into major, residential, pedestrian, and
tertiary classes and select up to ten spatially separated blocks per
city. Each selected block keeps its street name, nearest neighborhood
label, OSM way geometry, centroid, and driving side.

The street and neighborhood names describe a block in text and are used
in the prompts. The geometry defines the area where reference images are
collected, and the image panel records the block's observed appearance.
We truncate ways longer than 2\,km and require block centroids to be at
least 400\,m apart, so that each block remains a local unit that can be
compared with other streets in the same city.

We query Mapillary~\citep{mapillary} along each block geometry and retain
images that satisfy fixed camera, daylight, spatial, scene-content, and
image-quality filters. Appendix~\ref{app:curation} describes the filters
and their thresholds.
The reference index contains \num{\NUMIMAGESVTHREE} block--image
assignments for \num{\NUMBLOCKS} blocks. Because nearby blocks can share
images, these assignments cover \num{7433} distinct Mapillary images.
Table~\ref{tab:benchmark_stats} summarizes the benchmark.

\begin{table}[htbp]
\centering
\caption{GeoFidelity-Bench summary.}
\label{tab:benchmark_stats}
\TableStyle
\begin{tabular}{lrl}
\toprule
Statistic & Value & Notes \\
\midrule
Cities                     & \NUMCITIESVTHREE  & six continents \\
Named street blocks        & \NUMBLOCKS        & OSM ways, four road strata \\
Reference assignments      & \num{\NUMIMAGESVTHREE} & \num{7433} distinct images \\
Average images / block     & \NUMAVGBLOCK      & min 25 per block \\
Prompt levels              & 3                 & city; + street, neighborhood; + GPS \\
Generator models           & 6                 & open-weight T2I models \\
Retrieval entries / block  & up to 5           & target + available hard negatives \\
\bottomrule
\end{tabular}
\end{table}

\subsection{Reference Panels}

Each block has at least 25 reference images, with an average of
\num{\NUMAVGBLOCK}. A panel includes variation in camera heading,
lighting, and visible street features. We hold out four images from each
block as real-image queries and use the remaining images as its
evaluation reference panel. These real queries give a reference point
for generated images. Appendix~\ref{app:reference_structure}
describes visual variation across the city-level reference embeddings.

\subsection{Prompts and Retrieval Gallery}

We use three prompt levels. \textbf{L0} uses city and country only.
\textbf{L1} adds the street and neighborhood names; we also call L1
prompts \emph{named-block prompts}. \textbf{L2} appends the block
centroid to the L1 prompt as latitude and longitude written in text.
Each generator produces four images per block and prompt level. All templates use the same suffix
specifying photorealistic rendering, daytime, and clear weather, with no
additional style adjectives. Table~\ref{tab:prompt_templates} gives the
prompt templates.

\begin{table}[htbp]
\centering
\caption{Prompt templates for the prompt-conditioning study. Braces mark
metadata filled from each target block; all prompts use the same
photorealistic daytime suffix.}
\label{tab:prompt_templates}
\TableStyle
\begin{tabularx}{\linewidth}{@{}llX@{}}
\toprule
Level & Information & Template fragment \\
\midrule
L0 & city + country
   & \emph{taken in \{city\}, \{country\}} \\
L1 & L0 + street + neighborhood
   & \emph{taken on \{street\} in the \{neighborhood\} district of
      \{city\}, \{country\}} \\
L2 & L1 + raw GPS
   & \emph{taken on \{street\} in the \{neighborhood\} district of
      \{city\}, \{country\}, near GPS coordinates (\{lat\}, \{lon\})} \\
\bottomrule
\end{tabularx}
\end{table}

\paragraph{Retrieval gallery.}
\label{sec:hard_negatives}
For each target block, the retrieval gallery contains the target panel
and up to four negative panels: two other blocks in the same city, a
block in another city with the same driving side, and a block in a
random different city. The same-city negatives test whether an image
matches its target street better than plausible streets nearby, while
the other-city negatives test coarser geographic appearance. The negative
blocks of each target are fixed and shared by all models and prompt
conditions. Appendix~\ref{app:gen_configs} specifies how the gallery
handles missing and repeated entries.

\section{Metrics}
\label{sec:metrics}

Let $\mathcal{X}_g$ be the generated images for one block and
$\mathcal{X}_r$ be its real reference panel. We use panel cosine similarity
(CosSim), Diversity-Calibrated Set Fidelity (DCSF), Geo-Attribute Agreement
Score (GAAS), and retrieval accuracy. We also report the Maximum Mean
Discrepancy (MMD) component of DCSF. Table~\ref{tab:metric_roles} summarizes
how these quantities measure visual appearance, set distributions,
semantic composition, and discrimination among candidate locations.

\begin{table}[htbp]
\centering
\caption{Metric roles in GeoFidelity-Bench. CosSim is the main score;
the other reported quantities diagnose complementary errors.}
\label{tab:metric_roles}
\TableStyle
\begin{tabularx}{\linewidth}{@{}llXX@{}}
\toprule
Metric & Better & Main question & Typical failure exposed \\
\midrule
CosSim & high & visual match to the target panel & wrong local appearance \\
MMD & low & set distribution distance & distribution drift \\
DCSF & low & set distribution match & mode collapse or drift \\
GAAS & low & semantic attribute agreement & wrong road/sky/building mix \\
Retrieval & high & local discrimination & confusion with hard negatives \\
\bottomrule
\end{tabularx}
\end{table}

\paragraph{CosSim.}
CosSim is the primary metric. We encode all images with DINOv2
ViT-B/14~\citep{oquab2023dinov2}, average embeddings inside each panel,
and compute cosine similarity:
\begin{equation}
\mathrm{CosSim}(\mathcal{X}_g,\mathcal{X}_r)=
\frac{\bar{\mathbf{z}}_g \cdot \bar{\mathbf{z}}_r}
{\|\bar{\mathbf{z}}_g\|\,\|\bar{\mathbf{z}}_r\|},
\end{equation}
where $\bar{\mathbf{z}}_g$ and $\bar{\mathbf{z}}_r$ are the mean
embeddings of the generated and reference panels. Averaging gives each
image equal weight within its panel, so CosSim describes the mean visual
agreement between the generated set and the target panel.

\paragraph{MMD.}
Maximum Mean Discrepancy compares the generated and reference embedding
sets with an RBF kernel~\citep{gretton2012kernel}. We use the unbiased
squared estimate:
\begin{equation}
\widehat{\mathrm{MMD}}^2 =
\frac{1}{n(n-1)}\sum_{i\neq j} k(\mathbf{x}_i,\mathbf{x}_j)
+\frac{1}{m(m-1)}\sum_{i\neq j} k(\mathbf{y}_i,\mathbf{y}_j)
-\frac{2}{nm}\sum_{i,j} k(\mathbf{x}_i,\mathbf{y}_j).
\label{eq:mmd}
\end{equation}
Here $\mathbf{x}_i$ and $\mathbf{y}_j$ are generated and reference
embeddings, $n$ and $m$ are their panel sizes, and
$k(\mathbf{x},\mathbf{y})=\exp(-\gamma\|\mathbf{x}-\mathbf{y}\|^2)$.
Lower values indicate closer embedding distributions.
Appendix~\ref{app:gen_configs} gives the kernel settings and
finite-sample handling.

\paragraph{DCSF.}
Diversity-Calibrated Set Fidelity combines the MMD term in DINOv2 space
with a penalty when generated samples are less diverse than the
reference panel:
\begin{equation}
\mathrm{DCSF}=\max(0,\widehat{\mathrm{MMD}}^2(\mathcal{X}_g,\mathcal{X}_r))
+0.5\max(0,\mathrm{Div}(\mathcal{X}_r)-\mathrm{Div}(\mathcal{X}_g)).
\end{equation}
Here $\mathrm{Div}$ is the mean cosine distance between distinct images
within a panel. Lower DCSF indicates closer set distributions with a
smaller shortfall in generated diversity.

\paragraph{GAAS.}
Geo-Attribute Agreement Score measures interpretable scene attributes.
We run Mask2Former~\citep{cheng2022mask2former} trained on Mapillary
Vistas~\citep{neuhold2017mapillary}. For each image, we measure the pixel
ratios of road, sidewalk, building, vegetation, sky, pole, sign, and
vehicle groups. Let $p_g^{(a)}$ and $p_r^{(a)}$ be normalized histograms
of the pixel ratios for group $a$ in the generated and reference panels.
GAAS averages their Jensen--Shannon distances:
\begin{equation}
\mathrm{GAAS}(\mathcal{X}_g,\mathcal{X}_r)
=\frac{1}{8}\sum_{a=1}^{8}
\sqrt{\mathrm{JS}_2\!\left(p_g^{(a)},p_r^{(a)}\right)}.
\end{equation}
Here $\mathrm{JS}_2$ is the Jensen--Shannon divergence with base-2
logarithms. Lower GAAS indicates more similar semantic composition.
Appendix~\ref{app:gen_configs} gives numerical implementation details.

\paragraph{Hard-negative retrieval.}
For each generated image, we rank the mean embeddings of the target
reference panel and its available negative panels by cosine similarity.
For block $b$, let $r_{b,i}$ be the target panel's rank for generated
image $i$. Its top-1 retrieval accuracy is
\begin{equation}
\mathrm{Acc}_b=\frac{1}{n}\sum_{i=1}^{n}\mathbf{1}[r_{b,i}=1],
\end{equation}
where $\mathbf{1}$ is the indicator function and $n=4$ in our experiments.
We average this accuracy over target blocks. A gallery contains at most
five panels; blocks with missing negatives have smaller galleries.

We also compute a local margin. For each candidate panel, we average the
cosine similarities between the individual generated images and the
panel's mean embedding. The local margin is this value for the target
panel minus the larger of the two values for the same-city negatives.
A positive margin means that the generated images are, on average, closer
to the target block than to either same-city alternative.

\section{Evaluation Setup}
\label{sec:setup}

We evaluate six open-weight text-to-image models: SDXL~\citep{podell2023sdxl},
Stable Diffusion 3.5 Large~\citep{stablediffusion35}, FLUX.1-dev,
FLUX.1-schnell~\citep{flux2024}, PixArt-$\Sigma$~\citep{chen2024pixartsigma},
and HunyuanDiT~\citep{li2024hunyuan}.

Each model contributes a four-image panel for each of the 112 target
blocks under each prompt condition. Across six models, this gives 672
model--block panels per condition. We compare the three prompt levels
and the same-city controls using these matched model--block units.
All models generate $1024\times1024$ images.
Appendix~\ref{app:gen_configs} gives the model-specific step counts,
guidance scales, and precision settings (Table~\ref{tab:gen_configs}),
the negative prompt, and the reference splits used for each comparison.

\paragraph{Statistical comparisons.}
Prompt and control comparisons are paired by model and block, and their
95\% intervals use 5,000 bootstrap resamples of the matched pairs.
City-balanced estimates first average within each city and then give all
cities equal weight. Appendix~\ref{app:uncertainty} reports additional
city-level analyses.

\section{Results}
\label{sec:results}

\subsection{Reference-Panel Validation}
\label{sec:validation}

We first check whether the reference panels contain block-specific
appearance that a query can recover.
Table~\ref{tab:reference_hierarchy_main} compares a query panel of each
target block with the panels in its retrieval gallery: the same block,
the two same-city alternatives, a block in another city with the same
driving side, and a block in a random city. Held-out Real uses the four
real images held out from each target block as queries; the
six-generator mean uses generated panels under L1 prompts.

\begin{table}[htbp]
\centering
\caption{Reference hierarchy for real and generated queries. Values are mean DINOv2 image-to-panel
similarities to candidate panels, averaged over blocks where that
candidate type exists. Higher is more similar.}
\label{tab:reference_hierarchy_main}
\TableStyle
\begin{tabular}{lccccc}
\toprule
Query set & Same block & Same city (1) & Same city (2) & Same drive & Random \\
\midrule
Held-out Real   & 0.734 & 0.557 & 0.526 & 0.193 & 0.204 \\
Six-generator mean & 0.451 & 0.449 & 0.426 & 0.143 & 0.150 \\
\bottomrule
\end{tabular}
\end{table}

Held-out real images are clearly closer to their own block than to the
same-city alternatives (0.734 versus 0.557 and 0.526), so the panels
contain block-specific appearance. For generated images, similarity to
the target panel (0.451) is nearly equal to that to the first same-city
alternative (0.449) and only slightly higher than that to the second
(0.426). Both query types are much closer to same-city panels than to
panels from other cities, while generated images show much weaker
separation between the target block and same-city alternatives.

\subsection{Prompt Conditioning}

We next test how the location information in the prompt affects
similarity to the target panel. Table~\ref{tab:prompt_ablation} reports
CosSim for each model under city-only prompts (L0), street and
neighborhood names (L1), and additional GPS coordinates (L2).

\begin{table}[htbp]
\centering
\caption{CosSim under prompt-conditioning levels. Higher is better.}
\label{tab:prompt_ablation}
\TableStyle
\begin{tabular}{lccc}
\toprule
Model & L0 & L1 & L2 \\
\midrule
SDXL             & \VThreeLZeroSDXLC & \VThreeLOneSDXLC & \VThreeLTwoSDXLC \\
SD~3.5 Large     & \VThreeLZeroSDLC  & \VThreeLOneSDLC  & \VThreeLTwoSDLC  \\
FLUX.1-dev       & \VThreeLZeroFLDC  & \VThreeLOneFLDC  & \VThreeLTwoFLDC  \\
FLUX.1-schnell   & \VThreeLZeroFLSC  & \VThreeLOneFLSC  & \VThreeLTwoFLSC  \\
PixArt-$\Sigma$  & \VThreeLZeroPXC   & \VThreeLOnePXC   & \VThreeLTwoPXC   \\
HunyuanDiT       & \VThreeLZeroHDC   & \VThreeLOneHDC   & \VThreeLTwoHDC   \\
\bottomrule
\end{tabular}
\end{table}

Street and neighborhood names improve CosSim for every model
(Table~\ref{tab:prompt_ablation}). Moving from L0 to L1 raises mean
CosSim by +\VThreeDeltaLOneLZeroC{} over 672 matched (model, block)
pairs, with a bootstrap 95\% interval of [0.029, 0.042]
(Table~\ref{tab:prompt_uncertainty_main}). Giving each city equal weight
yields a similar gain of +0.042, and 21 of 25 cities show a positive
change. Adding raw GPS coordinates to L1 changes mean CosSim by only
+\VThreeDeltaLTwoLOneC{} (+0.006 with equal city weights), and only 14 of
25 cities show a positive change.

\begin{table}[htbp]
\centering
\caption{Prompt-ablation uncertainty checks for CosSim. Matched units
are \texttt{(model, block)} pairs; city-balanced units give each city
equal weight.}
\label{tab:prompt_uncertainty_main}
\TableStyle
\begin{tabular}{lcccc}
\toprule
Comparison & \shortstack{Matched delta\\(95\% CI)} & Positives & \shortstack{City-balanced delta\\(95\% CI)} & \shortstack{Cities with\\positive delta} \\
\midrule
L1 $-$ L0 & +0.036 [0.029, 0.042] & 438 / 672 & +0.042 [0.025, 0.060] & 21 / 25 \\
L2 $-$ L1 & +0.004 [0.001, 0.008] & 382 / 672 & +0.006 [0.001, 0.012] & 14 / 25 \\
\bottomrule
\end{tabular}
\end{table}

\subsection{Prompt-Specificity Controls}
\label{sec:prompt_controls}

The L1 gain could come from the correct names or from adding any
street-level description to the prompt. To separate the two, we keep the
L1 template, city, and country, and replace the street name, the
neighborhood label, or both with those of another benchmark block,
preferring blocks from the same city. These control prompts have the
same structure as L1 but point to the wrong location. Each control is
compared with L1 images evaluated against the same reference panels.
Table~\ref{tab:prompt_controls} reports the paired results.

\begin{table}[htbp]
\centering
\caption{Six-model prompt-specificity controls. Benefits are paired
L1-minus-control differences over 672 matched (model, block) pairs.}
\label{tab:prompt_controls}
\TableStyle
\begin{tabularx}{\linewidth}{@{}Xcccc@{}}
\toprule
Control prompt & CosSim & \shortstack{CosSim benefit\\(95\% CI)} & Retrieval & \shortstack{Retrieval benefit\\(95\% CI)} \\
\midrule
Wrong street & 0.498 & +0.001 [$-$0.004, 0.006] & 0.373 & +0.025 [0.008, 0.042] \\
Shuffled neighborhood & 0.492 & +0.007 [0.003, 0.012] & 0.361 & +0.038 [0.020, 0.057] \\
Wrong street and neighborhood & 0.488 & +0.011 [0.005, 0.017] & 0.344 & +0.054 [0.034, 0.074] \\
\bottomrule
\end{tabularx}
\end{table}

Retrieval accuracy is lower under all three controls than under L1, with
the largest decrease when both names are replaced
(Table~\ref{tab:prompt_controls}). The correct names therefore carry
block-specific information beyond the prompt template. Replacing only the
street name leaves CosSim almost unchanged (+0.001), whereas shuffling the
neighborhood label or replacing both names lowers it by 0.007 and 0.011.

\subsection{Text Alignment and Geographic Fidelity}
\label{sec:text_alignment}
Text-alignment scores are a common check that a generated image depicts
its prompt. We test whether they also indicate block-level
fidelity by comparing the geographic metrics with CLIP
ViT-B/32~\citep{radford2021clip} image--text cosine similarity. We use two
texts: the exact generation prompt and a location-only template, ``A
photo of \{street\} in the \{neighborhood\} district of \{city\},
\{country\}.'' The latter retains the location names and omits the
photographic instructions. For each text, we average the four image--text
scores within each of the 672 L1 panels.

Table~\ref{tab:clip_comparison} reports Spearman correlations with target
CosSim, the local margin, and retrieval accuracy. The area under the
receiver operating characteristic curve (AUROC) measures how well CLIP
predicts whether all four images retrieve the target;
0.5 corresponds to chance discrimination. The 95\% intervals use
5,000 bootstrap resamples of blocks, keeping all six models together
within each sampled block.

\begin{table}[!htbp]
\centering
\caption{Association between CLIP text alignment and geographic fidelity
across 672 L1 panels. Brackets give 95\% block-bootstrap intervals.
The first three rows report Spearman $\rho$; the last reports AUROC for
all four images retrieving the target.}
\label{tab:clip_comparison}
\TableStyle
\begin{tabular}{lcc}
\toprule
Geographic outcome & Exact prompt & Location-only text \\
\midrule
Target CosSim & 0.112 [$-0.008$, 0.231] & 0.238 [0.114, 0.353] \\
Local margin & 0.005 [$-0.117$, 0.126] & 0.123 [$-0.012$, 0.249] \\
Retrieval accuracy & 0.041 [$-0.068$, 0.149] & 0.105 [$-0.019$, 0.222] \\
\midrule
All-four retrieval AUROC & 0.508 [0.442, 0.572] & 0.579 [0.505, 0.652] \\
\bottomrule
\end{tabular}
\end{table}

Exact-prompt alignment has weak associations with all three geographic
outcomes and near-chance discrimination of all-four retrieval success.
Location-only text has a higher correlation with target CosSim, but its
correlations with local margin and retrieval remain small, with intervals
that include zero. High alignment also does not prevent retrieval
failure: among the quarter of panels with the highest exact-prompt CLIP
scores, 46.4\% contain no image that retrieves the target block, compared
with 47.3\% over all 672 panels (Appendix~\ref{app:clip}).

We next ask whether CLIP scores respond to the local names themselves.
Unlike the controls in Section~\ref{sec:prompt_controls}, we hold the
generated L1 images fixed and change only the scoring text, replacing the
street name, the neighborhood label, or both, while retaining the city
and country. Table~\ref{tab:clip_name_changes} reports the
original-minus-altered CLIP cosine similarity, averaged over the four
images in each panel.

\begin{table}[!htbp]
\centering
\caption{CLIP sensitivity to local names with generated images held fixed.
Differences are original-minus-altered text cosine similarities over
672 L1 panels. Intervals use 5,000 block-bootstrap resamples with all
six models kept together.}
\label{tab:clip_name_changes}
\TableStyle
\begin{tabular}{lccc}
\toprule
Altered text & Mean difference & 95\% CI & \shortstack{Panels with\\positive difference} \\
\midrule
Wrong street & 0.0018 & [0.0009, 0.0027] & 60.1\% \\
Shuffled neighborhood & 0.0021 & [0.0010, 0.0033] & 58.0\% \\
Both local names & 0.0047 & [0.0030, 0.0064] & 64.6\% \\
\bottomrule
\end{tabular}
\end{table}

Correct names yield higher mean CLIP scores, with the largest difference
when both names are replaced. The differences are small, at most 0.0047,
and the altered text scores at least as high as the correct text in 35\%
to 42\% of panels. CLIP therefore responds only weakly to which street
and neighborhood an image is supposed to show.

\subsection{Model Comparison under Named-Block Prompts}

We compare the six generators under L1 and use real-image anchors to
interpret the score differences (Table~\ref{tab:l1_leaderboard}). Besides
Held-out Real, Random-Same-Country and Random-Global are panels of four
real reference images sampled from other blocks in the same country and
from all other blocks, respectively.

\begin{table}[H]
\centering
\caption{Six-generator L1 results with real-image anchors.
CosSim and retrieval are higher-is-better; DCSF, MMD, and GAAS are
lower-is-better. Bold marks the best generator for each metric.}
\label{tab:l1_leaderboard}
\TableStyle
\begin{tabular}{lccccc}
\toprule
Model & CosSim $\uparrow$ & DCSF $\downarrow$ & MMD $\downarrow$ & GAAS $\downarrow$ & Retrieval $\uparrow$ \\
\midrule
SDXL & \textbf{0.536} & \textbf{0.547} & \textbf{0.457} & \textbf{0.380} & 0.408 \\
SD~3.5 Large & 0.515 & 0.565 & 0.470 & 0.387 & 0.375 \\
FLUX.1-schnell & 0.499 & 0.571 & 0.481 & 0.395 & 0.382 \\
PixArt-$\Sigma$ & 0.486 & 0.625 & 0.522 & 0.401 & \textbf{0.422} \\
FLUX.1-dev & 0.484 & 0.634 & 0.527 & 0.402 & 0.388 \\
HunyuanDiT & 0.477 & 0.637 & 0.531 & 0.421 & 0.420 \\
\midrule
Six-model mean & 0.499 & 0.596 & 0.498 & 0.398 & 0.399 \\
Held-out Real & 0.902 & 0.036 & 0.020 & 0.285 & 0.855 \\
Random-Same-Country & 0.710 & 0.160 & 0.151 & 0.365 & 0.259 \\
Random-Global & 0.333 & 0.249 & 0.249 & 0.370 & 0.163 \\
\bottomrule
\end{tabular}
\end{table}

Under named-block prompts, SDXL has the best generator scores on
CosSim, DCSF, MMD, and GAAS (Table~\ref{tab:l1_leaderboard}).
PixArt-$\Sigma$ has the highest generator retrieval accuracy.
Held-out Real scores better than all six generators on every metric,
with retrieval accuracy of 0.855 compared with 0.399 for the
six-generator mean.

The anchors also show that CosSim and retrieval measure different
properties. Random-Same-Country reaches CosSim 0.710, higher than every
generator, whereas Random-Global reaches only 0.333. Real images from the
same country thus share much of the target panel's appearance even when
they show other streets. Because they come from other blocks, however,
their retrieval accuracy is only 0.259, lower than that of every
generator.

\subsection{Semantic and Qualitative Comparisons}
\label{sec:semantic_composition}

Moving from L0 to L1 changes GAAS by $-0.0033$ on average over all eight
semantic groups (95\% CI $[-0.0075,0.0006]$), where lower GAAS is better.
Restricted to the four large scene regions (road, building, vegetation,
and sky), the change is $-0.0089$ (95\% CI $[-0.0176,-0.0006]$).
Street and neighborhood names thus move the broad scene composition
slightly closer to the reference panels. Appendix~\ref{app:seg_audit}
reports sensitivity to the segmentation model.

\begin{figure}[tbp]
\centering
\includegraphics[width=\textwidth]{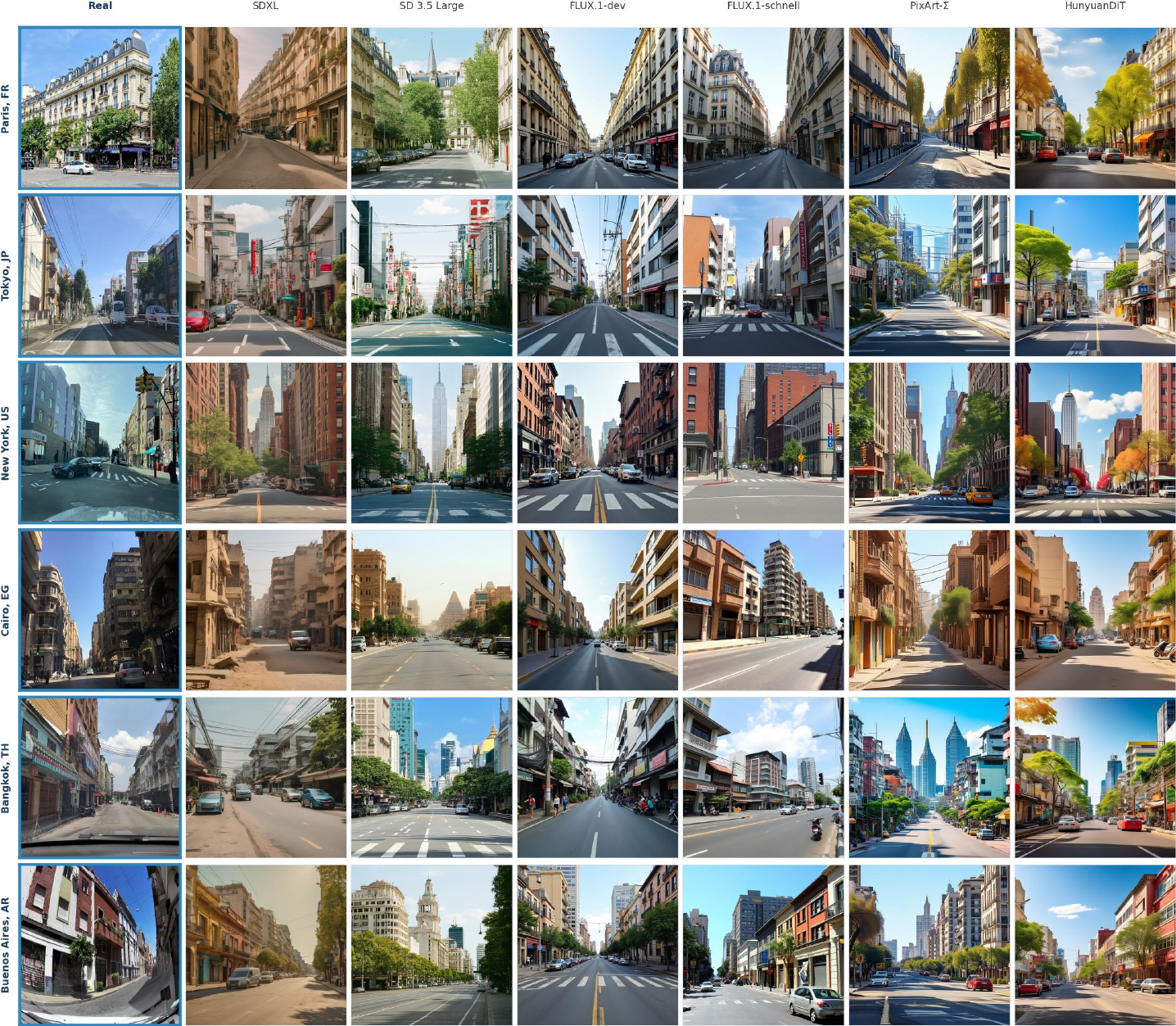}
\caption{Qualitative comparison between real reference images and six
open-weight generators under city-only prompts (L0). Generated images
often capture broad city style but differ from the references in curb
geometry, lane markings, storefronts, and the balance of road and sidewalk.}
\label{fig:qualitative}
\end{figure}

Figure~\ref{fig:qualitative} shows generations under city-only prompts
(L0). These prompts contain no block information, so the images show
what each model associates with a city. Generated images reproduce
tropical vegetation, European street widths, dense high-rise backgrounds,
or Tokyo-style sign density. They often miss details that make a block
recognizable: curb type, lane markings, storefront texture, pole
placement, and the balance between road, sidewalk, and facade.

\section{Limitations}
\label{sec:limitations}

Reference coverage follows Mapillary contribution patterns and OSM naming
density. Cities and blocks differ in their numbers of images, camera
headings, capture dates, and devices. Daylight and scene-content filters
further restrict the streetscapes represented. Appendix~\ref{app:uncertainty}
reports coverage counts, city-balanced estimates, and a leave-one-city-out
analysis of the main prompt effect.

The experiments supply street names and coordinates as text, so L2
measures whether text-only generators use coordinates written in the
prompt. Models that encode coordinates directly, such as that of
\citet{feng2025gps}, or that receive maps, satellite imagery,
trajectories, or retrieved images may use this information differently
and can be evaluated with the same reference panels and retrieval
galleries. A gain from street and neighborhood names may also reflect
appearance that a model associates with those names.

The metrics measure visual agreement with reference panels and do not
verify the state of a street at a particular time. Their values depend
on the DINOv2 encoder and the Mask2Former segmentation model used in
evaluation; Appendix~\ref{app:seg_audit} compares GAAS under a second
segmentation model.

Generated location images can reinforce place stereotypes or mislead
viewers when presented as photographic evidence. The benchmark uses
public Mapillary imagery with the platform's privacy processing, and its
intended use is evaluation of geographic fidelity. Appendix~\ref{app:bias_audit}
and the datasheet describe coverage biases and intended uses.

\section{Conclusion}
\label{sec:conclusion}

GeoFidelity-Bench evaluates generated street views against reference
panels from named OSM street blocks. It combines prompt ablations,
same-city controls, visual and semantic comparisons, and retrieval
against geographically related alternatives. Across six open-weight
generators, adding street and neighborhood names improves target-panel
similarity, and adding coordinates as text produces a smaller average
change. Generated images separate their target block from same-city
alternatives less reliably than held-out real images, with retrieval
accuracy of 0.399 versus 0.855, and CLIP text alignment does not predict
this separation.

The release contains \NUMBLOCKS{} blocks in \NUMCITIESVTHREE{} cities,
\num{\NUMIMAGESVTHREE} reference assignments, generated images, prompts, and
code. The same target panels and local alternatives can support future
evaluations of models conditioned on maps, coordinates, or retrieved
geographic imagery.

\paragraph{Data and Code Availability.}
The dataset (version \texttt{3.1.0}), generated images, evaluation code, and Croissant metadata
are publicly available at
\url{https://huggingface.co/datasets/moss-vector-714/GeoFidelity-Bench}.

\section*{Acknowledgments and Disclosures}
The author funded this research personally and received no external funding
or third-party support for this work. The author declares no competing
financial or other interests.

\clearpage
\bibliographystyle{plainnat}
\bibliography{references}

\newpage
\appendix
\section{Curation Pipeline Thresholds}
\label{app:curation}

Table~\ref{tab:tier_thresholds} collects the fixed controls used in the
block-level pipeline. Solar elevation follows the NOAA calculation
documentation~\citep{noaaSolar}; colorfulness uses the measure of
\citet{hasler2003colourfulness}. Thresholds are specified in the released
configuration files. The scene-content filters use a SigLIP urban-scene
classifier~\citep{zhai2023sigmoid} and semantic pixel ratios from
Mapillary-Vistas segmentation. Image-quality filters cover blur,
exposure, colorfulness, and aspect ratio.

\begin{table}[htbp]
\centering
\caption{Automatic reference-image filtering thresholds.}
\label{tab:tier_thresholds}
\TableStyle
\begin{tabularx}{\linewidth}{@{}lXX@{}}
\toprule
Stage & Threshold & Value \\
\midrule
Block/download & OSM way length             & $\geq 500$\,m (truncate at 2\,km) \\
Block/download & block spacing              & $\geq 400$\,m between centroids \\
Block/download & polyline resampling        & every 200\,m \\
Block/download & Mapillary bbox side        & 500\,m \\
Block/download & OSM buffer                 & \qty{30}{m} \\
Daylight/camera & sun elevation at capture         & $\geq 20^{\circ}$ \\
Daylight/camera & Mapillary \texttt{is\_pano}      & \texttt{false} \\
Daylight/camera & camera type                      & \texttt{perspective} \\
Urban scene & SigLIP urban prob (top-1)        & $\geq 0.35$ \\
Segmentation & sky ratio                        & $[0.01, 0.55]$ \\
Segmentation & building ratio                   & $[0.03, 0.70]$ \\
Segmentation & road ratio                       & $[0.01, 0.55]$ \\
Segmentation & vegetation ratio                 & $\leq 0.80$ \\
Segmentation & vehicle ratio                    & $\leq 0.40$ \\
Segmentation & urbanness floor                  & $\geq 0.05$ \\
Image quality & Laplacian variance               & $\geq 80$ \\
Image quality & mean luminance                   & $\in [55, 215]$ \\
Image quality & colorfulness                     & $\geq 14$ \\
Image quality & aspect ratio                     & $\in [0.5, 2.2]$ \\

\bottomrule
\end{tabularx}
\end{table}

\section{Generation and Evaluation Details}
\label{app:gen_configs}

Table~\ref{tab:gen_configs} gives the inference settings of each
generator: step count, guidance scale, output resolution, numerical
precision, and whether CPU offload is used. Each condition contains four
images per model and target block. L0 prompts contain only the city and
country, so the L0 images of a block are taken from a pool of city-only
generations for its city. The release includes the generated images for
all six prompt conditions (L0--L2 and the three controls), indexed by
model, target block, and prompt. Comparisons are paired by model and
target block.

\begin{table}[htbp]
\centering
\caption{Generator inference settings. Step counts, guidance scales, and
precision vary by model; output resolution is shared.}
\label{tab:gen_configs}
\TableStyle
\begin{tabular}{lccccc}
\toprule
Model & Steps & Guidance scale & Resolution & Precision & CPU offload \\
\midrule
SDXL               & 30 & 5.0 & $1024^2$ & fp16 & no  \\
SD~3.5 Large       & 28 & 3.5 & $1024^2$ & bf16 & yes \\
FLUX.1-dev         & 28 & 3.5 & $1024^2$ & bf16 & yes \\
FLUX.1-schnell     & 4  & 0.0 & $1024^2$ & bf16 & yes \\
PixArt-$\Sigma$    & 20 & 4.5 & $1024^2$ & fp16 & no  \\
HunyuanDiT         & 50 & 5.0 & $1024^2$ & fp16 & no  \\
\bottomrule
\end{tabular}
\end{table}

\paragraph{Reference splits and retrieval entries.}
The prompt-level comparisons and the L1 model comparison come from one
evaluation run. The prompt-specificity controls come from a separate run
that also scores L1, and each L1-minus-control difference pairs the same
model and target block within that run. The block index lists the
retrieval negatives of each target. Missing negatives are omitted, and
if two negative entries refer to the same block, both are kept.

Held-out real queries are removed from the panel of their own block.
Blocks that meet, for example at a street crossing, can share Mapillary
images, so a held-out query can also appear in the panel of a same-city
negative. When such a query is scored against that negative, it is
partly compared with itself, which lowers its chance of retrieving the
target; Held-out Real retrieval is therefore a conservative reference
point. For generated queries, the shared image moves that negative panel
slightly toward the target's appearance and makes it harder to separate
from the target. The target panels used for CosSim, DCSF, MMD, and GAAS
never contain held-out queries.

\paragraph{Metric implementation.}
The RBF parameter $\gamma$ is the reciprocal of the median
positive squared distance in the pooled generated and reference
embeddings. We report $\max(0,\widehat{\mathrm{MMD}}^2)$, truncating negative
finite-sample estimates to zero. GAAS uses 10 bins on $[0,1]$ for each
semantic group's pixel-ratio histogram and adds $10^{-8}$ to each bin
before normalization. The evaluation also reports mean reciprocal rank. We average the
reciprocal of the target panel's rank over the four individual queries
and then over blocks. For the strongest-negative margin, we first average
query-to-panel cosine similarities for each candidate panel. The margin
is the target's mean similarity minus the largest mean similarity among
all available negative panels.

\paragraph{Compute resources.}
The generation runs used a single NVIDIA RTX 4090 (24\,GB). SD~3.5 Large, FLUX.1-dev, and
FLUX.1-schnell use CPU offload because their memory use is higher at
$1024^2$ resolution. The evaluation scripts cache image embeddings and
reuse them across prompt levels and controls. Additional CLIP and segmentation analyses used an NVIDIA RTX 4060
Laptop GPU (8\,GB).

\begin{samepage}
\paragraph{Negative prompt.}
All generators except the two FLUX models, which do not support negative
prompts, receive the following negative prompt:
\begin{quote}\itshape
panorama, fisheye, black and white, illustration, painting, cartoon,
3d render, watermark, text overlay, blurry, low resolution, night,
people close up, indoor.
\end{quote}
\end{samepage}

\section{City Structure in the Reference Set}
\label{app:reference_structure}

\begin{figure}[tbp]
\centering
\includegraphics[width=\textwidth]{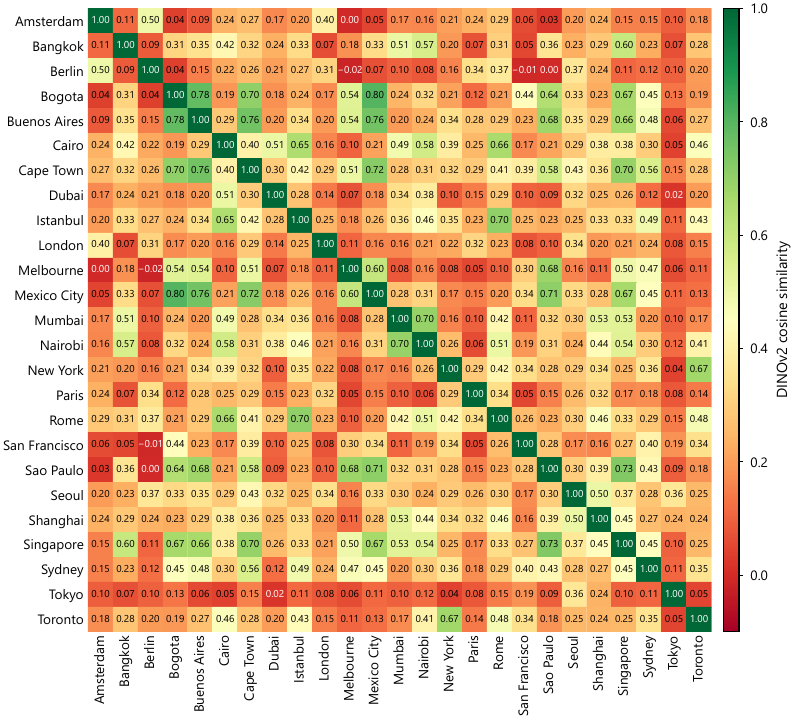}
\caption{Cross-city visual similarity matrix from real Mapillary
references. Each city is represented by the mean DINOv2 embedding of its retained
reference images; entries are cosine similarities between city means.}
\label{fig:cross_city}
\end{figure}

Figure~\ref{fig:cross_city} shows a $25{\times}25$ DINOv2 similarity
matrix over real Mapillary images aggregated by city. The off-diagonal
mean is \CROSSCITYMEAN{}. The diagonal is $1.000$ by construction.
Visual distance between cities increases weakly with geographic
distance: over the 300 city pairs, the Spearman correlation between
geodesic distance and visual distance is $\rho=\GEOVISSPEARMAN$
($p=\GEOVISPVAL$).

\section{Uncertainty and Sensitivity Analyses}
\label{app:uncertainty}

The paired and city-balanced prompt comparisons are reported in
Table~\ref{tab:prompt_uncertainty_main}. To measure the influence of
individual cities on the L1-minus-L0 result, we repeat the comparison 25
times, each time omitting all blocks from one city, and average the
matched differences over the remaining model--block pairs. The gain
remains positive in every repetition: CosSim ranges from +0.0303 to
+0.0381, retrieval accuracy from +0.0397 to +0.0542, and local margin
from +0.0131 to +0.0154.

Coverage is uneven: Tokyo contributes 12.0\% of all reference images,
New York 10.1\%, and the smallest cities retain only one block. The
city-balanced estimates give each city equal weight.

\section{Segmentation Sensitivity}
\label{app:seg_audit}

We first check for city-level segmentation failures, which would appear
as masks with unusually low label diversity or a single dominant label.
For each reference image, we compute the entropy and the largest pixel
share over 12 aggregated semantic groups. City medians are standardized
by the across-city median and $1.4826$ times the median absolute
deviation. No city has standardized entropy below $-2.5$ or
concentration above $+2.5$, so no city shows this failure pattern.

We then vary the semantic groups and the segmentation model. Restricting
GAAS to road, building, vegetation, and sky preserves all 15 pairwise L1
model orderings relative to the eight-group score (Spearman
$\rho=1.000$); Section~\ref{sec:semantic_composition} reports the
corresponding prompt-level differences. We also compute GAAS with
SegFormer-B5~\citep{xie2021segformer} trained on
Cityscapes~\citep{cordts2016cityscapes}, mapping its labels to the same
eight groups. Across 672 L1 panels, its scores have Spearman
$\rho=0.739$ with Mask2Former GAAS. The six-model rank correlation is
$0.714$, with 11 of 15 pairwise orderings preserved. SD~3.5 Large ranks
first with SegFormer, whereas SDXL ranks first with Mask2Former.

\section{Text-Alignment Comparison}
\label{app:clip}

Table~\ref{tab:clip_comparison} compares CLIP text alignment with the
geographic metrics. To examine the extremes, we split the 672 L1 panels
into quartiles of CLIP similarity and count panels in which no image
retrieves the target block. For exact-prompt alignment, this proportion
is 46.4\% in the highest quartile (168 panels) and 53.0\% in the lowest.
For location-only alignment, it is 44.0\% and 54.2\%. Over all panels,
it is 47.3\%.

\section{Benchmark City List}
\label{app:cities}

Table~\ref{tab:full_city_list} reports city coordinates, driving conventions,
and post-curation counts for the 25 cities. The retained cities range
from one to ten blocks, reflecting differences in available imagery.

\begin{table}[htbp]
\centering
\caption{Benchmark city-level statistics. Reference counts are block--image
assignments, including images assigned to more than one block.}
\label{tab:full_city_list}
\TableStyle
\begin{tabular}{llccccc}
\toprule
City & Country & Latitude & Longitude & \shortstack{Driving\\side} & Blocks & \shortstack{Reference\\assignments} \\
\midrule
\CITYTABLEVTHREE \\
\midrule
\textbf{Total} & & & & & \textbf{\NUMBLOCKS} & \textbf{\num{\NUMIMAGESVTHREE}} \\
\bottomrule
\end{tabular}
\end{table}

\section{Dataset Bias and Governance}
\label{app:bias_audit}

\paragraph{Coverage bias.}
The benchmark remains constrained by Mapillary's global contribution
pattern. Named-block carving depends on OSM naming density and
contributor activity. Informal
settlements, peri-urban fringes, and cities with sparse Mapillary use
remain under-represented. The current release spans six continents and
23 countries, but should not be read as geographically uniform.

\paragraph{Capture bias.}
The sun-elevation gate intentionally biases the benchmark toward
daytime imagery. Contributor capture setups are heterogeneous (handheld
phones, dashcams, and cameras mounted on bicycles or carried by
pedestrians), and we retain this variation except for panoramic and
fisheye imagery. As a result, the benchmark measures
geographic fidelity under a realistic street-photography mixture, not
under a single canonical camera rig.

\paragraph{Semantic-filter bias.}
Segmentation- and signage-sensitive filters may behave differently
across scripts, urban forms, and low-resource regions. We mitigate
this by using a globally-trained Mapillary-Vistas backbone and by
releasing the full pipeline so thresholds can be re-run or retuned for
new regions.

\section{Datasheet}

\paragraph{Motivation.}
GeoFidelity-Bench evaluates geographic fidelity in
location-conditioned street-view generation, filling a gap left by
quality-focused and diversity-focused benchmarks.

\paragraph{Composition.}
\num{\NUMIMAGESVTHREE} reference assignments covering \num{7433} distinct
street-level images from \NUMBLOCKS{} named
street blocks across \NUMCITIESVTHREE{} cities on six continents, sourced
from Mapillary (CC-BY-SA). The reference index includes GPS and available capture metadata.

\paragraph{Collection process.}
Blocks are carved from named OSM ways. Images are retrieved with
small bounding-box Mapillary queries along each block and filtered by the
automated stages in Table~\ref{tab:tier_thresholds}; the reference index
records the retained images.

\paragraph{Uses.}
Evaluating geographic fidelity of image generation models. Not
intended for surveillance, person identification, or any use that
could harm individuals depicted in Mapillary imagery.

\paragraph{Distribution.}
The dataset (version \texttt{3.1.0}), generated images, evaluation code,
and Croissant metadata are distributed through Hugging Face at
\url{https://huggingface.co/datasets/moss-vector-714/GeoFidelity-Bench}.
The reference images retain Mapillary's CC-BY-SA 4.0 terms. OSM-derived
metadata are subject to ODbL 1.0 and attribution to OpenStreetMap
contributors. Benchmark code is released under the MIT license.
Generated images retain the applicable source-model terms, listed in
the dataset card.

\newpage
\section*{NeurIPS Paper Checklist}

\begin{enumerate}

\item {\bf Claims}
    \item[] Question: Do the main claims made in the abstract and introduction accurately reflect the paper's contributions and scope?
    \item[] Answer: \answerYes{}
    \item[] Justification: Sections~\ref{sec:intro}--\ref{sec:results} describe the named-block benchmark, evaluation protocol, and results. The claims concern visual agreement with reference panels and discrimination among candidate locations.

\item {\bf Limitations}
    \item[] Question: Does the paper discuss the limitations of the work performed by the authors?
    \item[] Answer: \answerYes{}
    \item[] Justification: Section~\ref{sec:limitations} discusses reference coverage, text-based conditioning, representation dependence, and possible misuse. Appendices~\ref{app:uncertainty} and~\ref{app:bias_audit} describe geographic coverage.

\item {\bf Theory assumptions and proofs}
    \item[] Question: For each theoretical result, does the paper provide the full set of assumptions and a complete (and correct) proof?
    \item[] Answer: \answerNA{}
    \item[] Justification: The paper presents a benchmark and empirical analyses rather than theoretical results.

    \item {\bf Experimental result reproducibility}
    \item[] Question: Does the paper fully disclose all the information needed to reproduce the main experimental results of the paper to the extent that it affects the main claims and/or conclusions of the paper (regardless of whether the code and data are provided or not)?
    \item[] Answer: \answerYes{}
    \item[] Justification: Sections~\ref{sec:benchmark}--\ref{sec:setup} and Appendix~\ref{app:gen_configs} describe the benchmark, metrics, prompts, and inference settings. The public release provides images, evaluation code, intermediate outputs, and per-block scores with analysis scripts for reproducing the reported comparisons.

\item {\bf Open access to data and code}
    \item[] Question: Does the paper provide open access to the data and code, with sufficient instructions to faithfully reproduce the main experimental results, as described in supplemental material?
    \item[] Answer: \answerYes{}
    \item[] Justification: The reference and generated images, metadata, code, and intermediate results are publicly available in the linked Hugging Face repository. The code archive includes dependencies and commands for reproducing the main statistical analyses.

\item {\bf Experimental setting/details}
    \item[] Question: Does the paper specify all the training and test details (e.g., data splits, hyperparameters, how they were chosen, type of optimizer) necessary to understand the results?
    \item[] Answer: \answerYes{}
    \item[] Justification: Sections~\ref{sec:benchmark}--\ref{sec:setup} specify the prompt conditions, metric definitions, and evaluation protocol. Appendix~\ref{app:curation} gives the filtering thresholds, and Appendix~\ref{app:gen_configs} gives the inference settings and additional evaluation details.

\item {\bf Experiment statistical significance}
    \item[] Question: Does the paper report error bars suitably and correctly defined or other appropriate information about the statistical significance of the experiments?
    \item[] Answer: \answerYes{}
    \item[] Justification: The paper reports paired bootstrap intervals and city-balanced estimates. The resampling units and 95\% intervals are specified in Section~\ref{sec:setup} and Appendix~\ref{app:uncertainty}.

\item {\bf Experiments compute resources}
    \item[] Question: For each experiment, does the paper provide sufficient information on the computer resources (type of compute workers, memory, time of execution) needed to reproduce the experiments?
    \item[] Answer: \answerNo{}
    \item[] Justification: Appendix~\ref{app:gen_configs} reports GPU models, memory, CPU offload, and inference settings. It does not provide a complete accounting of execution time and total project compute.
    
\item {\bf Code of ethics}
    \item[] Question: Does the research conducted in the paper conform, in every respect, with the NeurIPS Code of Ethics \url{https://neurips.cc/public/EthicsGuidelines}?
    \item[] Answer: \answerYes{}
    \item[] Justification: The work uses public street-view imagery with source attribution and discusses privacy, geographic coverage, and possible misuse in Section~\ref{sec:limitations} and Appendix~\ref{app:bias_audit}.

\item {\bf Broader impacts}
    \item[] Question: Does the paper discuss both potential positive societal impacts and negative societal impacts of the work performed?
    \item[] Answer: \answerYes{}
    \item[] Justification: The benchmark supports evaluation of geographic errors in generated images. Section~\ref{sec:limitations} discusses inaccurate place depictions, stereotypes, and misuse as photographic evidence.
    
\item {\bf Safeguards}
    \item[] Question: Does the paper describe safeguards that have been put in place for responsible release of data or models that have a high risk for misuse (e.g., pre-trained language models, image generators, or scraped datasets)?
    \item[] Answer: \answerYes{}
    \item[] Justification: Section~\ref{sec:limitations} describes reliance on Mapillary privacy processing and the intended evaluation use. The release identifies real and generated images and retains source licenses and attribution.

\item {\bf Licenses for existing assets}
    \item[] Question: Are the creators or original owners of assets (e.g., code, data, models), used in the paper, properly credited and are the license and terms of use explicitly mentioned and properly respected?
    \item[] Answer: \answerYes{}
    \item[] Justification: The paper cites the image sources, geographic data, encoders, segmentation models, and generators. The datasheet and dataset card specify source licenses, attribution, and model-dependent terms for generated images.

\item {\bf New assets}
    \item[] Question: Are new assets introduced in the paper well documented and is the documentation provided alongside the assets?
    \item[] Answer: \answerYes{}
    \item[] Justification: The datasheet and public release document composition, collection, intended uses, source licenses, and maintenance.

\item {\bf Crowdsourcing and research with human subjects}
    \item[] Question: For crowdsourcing experiments and research with human subjects, does the paper include the full text of instructions given to participants and screenshots, if applicable, as well as details about compensation (if any)? 
    \item[] Answer: \answerNA{}
    \item[] Justification: The experiments reported in this paper use automated evaluation of images and do not include crowdsourcing or a human-participant study.

\item {\bf Institutional review board (IRB) approvals or equivalent for research with human subjects}
    \item[] Question: Does the paper describe potential risks incurred by study participants, whether such risks were disclosed to the subjects, and whether Institutional Review Board (IRB) approvals (or an equivalent approval/review based on the requirements of your country or institution) were obtained?
    \item[] Answer: \answerNA{}
    \item[] Justification: The reported experiments do not include a human-participant study or collect participant data.

\item {\bf Declaration of LLM usage}
    \item[] Question: Does the paper describe the usage of LLMs if it is an important, original, or non-standard component of the core methods in this research? Note that if the LLM is used only for writing, editing, or formatting purposes and does \emph{not} impact the core methodology, scientific rigor, or originality of the research, declaration is not required.
    \item[] Answer: \answerNA{}
    \item[] Justification: LLM assistance was used for writing and editing. It was not an important, original, or non-standard component of the core research method.

\end{enumerate}

\end{document}